\documentclass[runningheads]{llncs}

\usepackage[T1]{fontenc}
\usepackage{booktabs}
\usepackage{multirow}
\usepackage{makecell}
\usepackage{amsmath}
\usepackage{graphicx,verbatim}
\usepackage{tcolorbox}
\usepackage{xcolor}
\usepackage{amssymb}
\usepackage{subcaption}
\usepackage{hyperref}

\begin{document}

\title{Unleashing Video Language Models for Fine-grained HRCT Report Generation}

\begin{comment}  %% Removed for anonymized MICCAI 2025 submission
\author{First Author\inst{1}\orcidID{0000-1111-2222-3333} \and
Second Author\inst{2,3}\orcidID{1111-2222-3333-4444} \and
Third Author\inst{3}\orcidID{2222--3333-4444-5555}}
%
\authorrunning{F. Author et al.}
\institute{Princeton University, Princeton NJ 08544, USA \and
Springer Heidelberg, Tiergartenstr. 17, 69121 Heidelberg, Germany
\email{lncs@springer.com}\\
\url{http://www.springer.com/gp/computer-science/lncs} \and
ABC Institute, Rupert-Karls-University Heidelberg, Heidelberg, Germany\\
\email{\{abc,lncs\}@uni-heidelberg.de}}

\end{comment}

\author{Yingying Fang\inst{1}\textsuperscript{*} \and
Huichi Zhou\inst{1}\textsuperscript{*} \and
KinHei Lee\inst{1}\textsuperscript{*} \and
Yijia Wang\inst{1}\textsuperscript{*} \and
Zhenxuan Zhang\inst{1} \and
Jiahao Huang\inst{1} \and
Guang Yang\inst{1,2}
}
 %% Added for anonymized MICCAI 2025 submission
\authorrunning{Y. Fang, H. Zhou, K. Lee, Y. Wang et al.}
\institute{
Bioengineering Department, Imperial College London, London, UK \and
School of Biomedical Engineering \& lmaging Sciences, King's College London,
London,UK 
\\
\email{y.fang@imperial.ac.uk}
\\
}

\renewcommand{\thefootnote}{\fnsymbol{footnote}}
\footnotetext[1]{These authors contributed equally to this work.}

\maketitle

\begin{abstract}

Generating precise diagnostic reports from High-Resolution Computed Tomography (HRCT) is critical for clinical workflow, yet it remains a formidable challenge due to the high pathological diversity and spatial sparsity within 3D volumes. While Video Language Models (VideoLMs) have demonstrated remarkable spatio-temporal reasoning in general domains, their adaptability to domain-specific, high-volume medical interpretation remains underexplored.
In this work, we present AbSteering, an abnormality-centric framework that steers VideoLMs toward precise HRCT report generation. Specifically, AbSteering introduces: (i) an abnormality-centric Chain-of-Thought scheme that enforces abnormality reasoning, and (ii) a Direct Preference Optimization objective that utilizes clinically confusable abnormalities as hard negatives to enhance fine-grained discrimination. Our results demonstrate that general-purpose VideoLMs possess strong transferability to high-volume medical imaging when guided by this paradigm. Notably, AbSteering outperforms state-of-the-art domain-specific CT foundation models, which are pretrained with large-scale CTs, 
achieving superior detection sensitivity while simultaneously mitigating hallucinations.
Our data and model weights are released at \url{https://anonymous.4open.science/r/hrct-report-generation-video-vlm-728C/}.
\end{abstract}

\section{Introduction}

High-Resolution Computed Tomography (HRCT) is a key modality for the definitive diagnosis and longitudinal monitoring of thoracic and cardiopulmonary diseases, providing volumetric scans with rich texture and fine anatomical detail. In parallel, AI-driven medical report generation has become an active research direction, as it can reduce clinical workload, standardize diagnostic narratives, and mitigate inter-observer variability. While report generation for 2D chest X-rays has progressed rapidly, extending these advances to 3D HRCT remains substantially more challenging. Compared with projection imaging, HRCT introduces both (i) prohibitive computational and memory costs due to hundreds of slices per study, and (ii) a more difficult visual understanding problem, where clinically critical abnormalities are subtle, spatially localized, and diverse, appearing sparsely across the volume and often being overwhelmed by dominant normal anatomical patterns.

Early attempts at CT report generation built on X-ray paradigms by compressing CT volumes into lower-dimensional representations and then reusing X-ray–oriented report generators~\cite{hamamci2024developing,tang2024work,cao2024bootstrapping}. With the emergence of large language models (LLMs), subsequent work (e.g., Dia-LLaMA~\cite{chen2025dia}) designed CT-specific visual encoders and connected them to LLM decoders to improve language quality and clinical style. More recently, modality-specific radiology foundation models have been explored to better handle 3D medical inputs and support report generation, such as RadFM~\cite{wu2023towards}, CT-CHAT~\cite{hamamci2026generalist} and M3D~\cite{bai2024m3d}. Building on these backbones, a line of work further focuses on strengthening CT feature extraction and representation learning based on the foundation models~\cite{chen20243d,li2025mu,deng2025mvketr}. Despite steady progress, most existing approaches still rely on training or heavily tailoring modality-specific encoders, which can be data- and compute-intensive, and they remain limited in fine-grained recognition of long-tail, localized abnormalities that drive report correctness.

In a different vein, recent advances in \textbf{video-language models (VideoLMs)} have significantly improved their ability to reason over long visual sequences by modeling temporal and spatial dependencies~\cite{lin2023video,wang2024emu3,zhang2024vision,wang2024qwen2,zhu2025internvl3}.
These VideoLMs are trained to align dynamic visual content with language, enabling strong performance on captioning and multi-step video reasoning. Importantly, an HRCT volume can be naturally viewed as a video-like slice sequence, which makes VideoLMs a promising candidate for volumetric report generation. 
However, despite their strong performance, VideoLMs are typically pretrained on natural video data and therefore lack domain-specific knowledge---especially for medical 3D imaging where recognizing subtle and localized abnormalities requires specialized clinical understanding. This mismatch raises three key questions: (1) \textit{Can the encoders in VideoLMs capture clinically relevant 3D features and support accurate CT report generation?} (2) \textit{How can we adapt general-domain VideoLMs to domain-specific medical reporting in a highly efficient manner?} (3) \textit{How does such transfer compare with modality-specific CT foundation models in terms of report accuracy and clinical fidelity?}.

To address these questions, we investigate the transferability of pretrained VideoLMs to high-volume radiology interpretation and propose AbSteering, an abnormality-centric framework that steers VideoLMs toward precise radiology interpretation by tructured reasoning pathways and discriminative preference optimization. Concretely, our framework introduces:
(i) an \textbf{abnormality-centric Chain-of-Thought (CoT) training} scheme built on curated reports with structured templates. This explicitly prompts clinically critical reasoning before final report generation, constraining the model to learn diverse disease categories while suppressing hallucinated or normal-tissue-dominated descriptions.
(ii) a \textbf{fine-grained abnormality discrimination objective based on Direct Preference Optimization (DPO)}, which introduces clinically confusable abnormalities within the same anatomical region as hard negatives to better resolve subtle inter-abnormality differences and reduce hallucinations.

Our contributions are three-fold: (1) \textbf{Cross-modal transferability.} We systematically study the transferability of general-domain VideoLMs and show that, under limited data, they transfer effectively to 3D medical imaging, providing an efficient alternative to training modality-specific foundation models from scratch; (2) \textbf{Method.} We propose an efficient abnormality-centric language-steering framework that successfully adapts VideoLMs and achieves state-of-the-art performance on fine-grained clinical efficacy metrics; and (3) \textbf{Dataset.} We curate \textbf{CT-RATE-AB}, a dataset designed to facilitate CoT training in medical imaging and to evaluate diversity and fine-grained abnormality recognition for high-volume chest CT understanding.

\section{Method}
\begin{figure}[t]
    \centering

    \includegraphics[width=\linewidth]{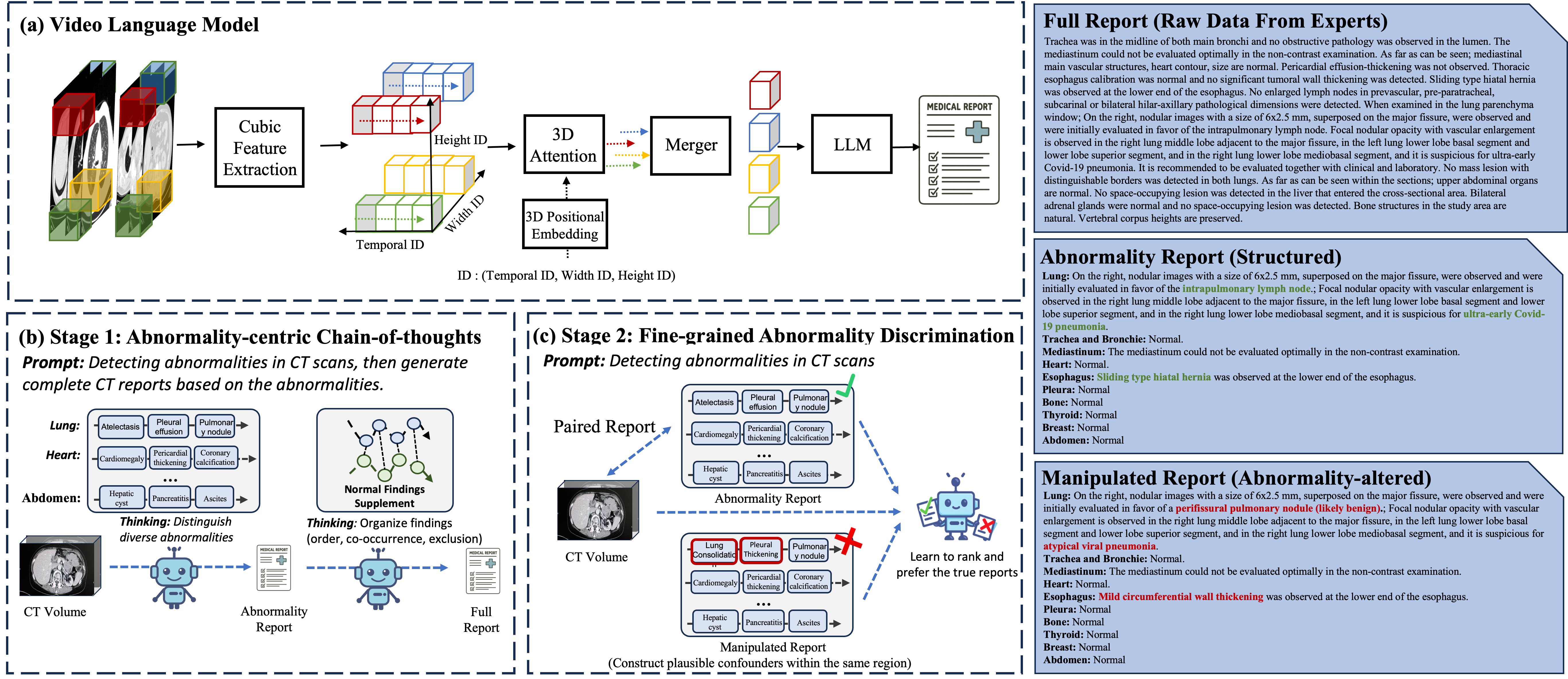}
    
    \caption{\textbf{Left:} overall framework for adapting current VideoLMs with the proposed AbSteering in two stages: (a) the typical backbone of current VideoLMs; (b) Stage 1, abnormality-centric CoT training; and (c) Stage 2, fine-grained abnormality discrimination via DPO. \textbf{Right:} examples of abnormality-centric and manipulated report samples.}
    \label{fig:framework}
\end{figure}

As illustrated in Fig.~\ref{fig:framework}, our framework leverages a pretrained VideoLM backbone as its foundation. The AbSteering method is implemented in two stages:

\textbf{Stage 1: abnormality-centric CoT training.} We reformulate the vision-to-text task into a structured sequence that transitions from abnormality detection to report composition. 
This decomposition compels the model to explicitly reason over pathological findings, while simultaneously learning the underlying clinical correlations and standardized report structures prior to final generation.

\textbf{Stage 2: fine-grained abnormality discrimination via DPO.} To enhance the model's ability to distinguish between subtle pathological nuances and mitigate hallucinations, we introduce a preference optimization strategy. By contrasting clinically similar abnormalities within the same anatomical region, this stage forces the model to resolve fine-grained diagnostic features through hard-negative discriminative learning.

\subsection{Architecture of Video-Language Models}

Most mainstream VideoLMs follow a common visual-language pipeline for processing long visual sequences. Given an input video $X \in \mathbb{R}^{T \times H \times W \times C}$, where $T$ denotes the temporal dimension and $(H,W)$ denote the spatial dimensions, the visual backbone first partitions the input into spatiotemporal cubes (or frame patches) and encodes them into visual tokens, $z^{(0)}_{t,i,j} = E_{\mathrm{cube}}(X_{t,i,j})$. These tokens are then processed by a spatiotemporal transformer with 3D attention, where positional information is injected through factorized 3D positional embeddings along the temporal, height, and width axes:
\begin{equation}
Z^{(l+1)} = \mathrm{Transformer}_{3D}\big(Z^{(l)};\, e_t^{\mathrm{temp}}, e_i^{\mathrm{h}}, e_j^{\mathrm{w}}\big), \qquad Z^{(l)}=\{z^{(l)}_{t,i,j}\}.
\end{equation}

After several layers, a merger (or projector/resampler) compresses the dense visual tokens into a smaller set of language-aligned tokens, $V = G(Z^{(L)})$, which are fed into a LLM for autoregressive generation.

From an architectural perspective, current VideoLMs and modality-specific CT foundation models are largely similar. Both typically follow the same high-level pipeline: volumetric or spatiotemporal tokenization, positional encoding, 3D attention-based visual encoding, token merging, and language decoding for report generation. 

The main difference therefore does not lie in the backbone structure itself, but in the training domain and supervision. VideoLMs are pretrained on natural videos, where the model learns generic temporal dynamics and broad visual-language alignment, whereas modality-specific CT foundation models are trained directly on medical volumes with domain-specific objectives and clinical reports. 

While CT-specific foundations incorporate stronger medical priors, VideoLMs provide a strong generic spatiotemporal backbone. This also motivates our study: if the architectural form is already well aligned with volumetric CT analysis, then the key challenge becomes how to efficiently adapt general-domain VideoLMs to domain-specific HRCT report generation.

\subsection{Abnormality-centric chain-of-thoughts}

To enable abnormality-centric reasoning, we normalize CT-RATE reports into a unified \texttt{(region: abnormality)} template. Using this structured-to-raw pairing, we implement a CoT process that decouples the task: the model first distills diagnostic findings as a reasoning anchor before synthesizing the final report.

\noindent\textbf{Structuring reports.}
To reorganize raw reports, we require the structured format to be both information-preserving and standardized across patients. Following the anatomical taxonomy in~\cite{zhang2024radgenome}, we group report content into ten regions: Lung, Trachea and Bronchi, Mediastinum, Heart, Esophagus, Pleura, Bone, Thyroid, Breast, and Abdomen. GPT-4o~\cite{achiam2023gpt} is used to assign each sentence to these regions, while abnormalities outside the taxonomy are categorized as "Others".

For verification, two auxiliary categories, uncategorized and repetitive, are introduced. Potential issues are identified through sentence-level matching and manually refined, resulting in the curated \textbf{\textit{CT-RATE-AB}} dataset, as exemplified in Fig.~\ref{fig:framework}.

\noindent\textbf{CoT training.} Given the curated abnormality tokens $R_{AB}$ and the full report $R_{Full}$, we optimize the model via a sequential generation objective. The training prompt explicitly instructs the model to detect abnormalities before composing the narrative. The construction of the target sequence is defined as:
\begin{equation}
\mathcal{L}_{gen} = -\sum_{t=1}^{T} \log P(y_t \mid x, y_{<t})
\end{equation}
% $$\mathcal{L}_{gen} = -\sum_{t=1}^{T} \log P(y_t \mid x, y_{<t})$$
where the target $Y = [R_{AB}; R_{Full}]$. By enforcing this causal chain, the model is compelled to prioritize clinically significant features and subtle morphological distinctions. At the reasoning level, transitioning from discrete findings to a narrative allows the model to capture anatomical constraints and pathological inter-dependencies, such as the co-occurrence of related diseases or the mutual exclusivity of contradictory findings.

\subsection{Fine-grained abnormality discrimination}

CT abnormalities often exhibit subtle and visually confusable patterns, making fine-grained discrimination highly dependent on domain-specific clinical knowledge. To improve the model’s ability to distinguish these subtle distinctions, we introduce a fine-grained preference learning strategy in Stage 2 that constructs fake reports by replacing true abnormalities with clinically confusable alternatives. We then explicitly teach the model to prefer the clinically correct report over the corrupted one through DPO. By learning which report is more faithful, the model is encouraged to attend to the subtle visual cues that determine report quality, thereby improving both fine-grained abnormality discrimination and hallucination suppression.

\noindent\textbf{Construction of $R_{\text{AB\_Fake}}$.}
We construct $R_{\text{AB\_Fake}}$ automatically from $R_{\text{AB}}$ using GPT-4o. For each abnormality-centric report, GPT-4o is instructed to replace the target abnormality with an anatomically matched but clinically distinct abnormality that could plausibly occur in the same region, while keeping the original region label, sentence template, and all other positional information unchanged. This yields hard negative reports that preserve fluency and structural consistency but alter the clinically critical abnormality semantics, thereby providing an effective contrastive signal for preference optimization.

\noindent\textbf{Direct Preference Optimization.} We further refine the Stage 1 model (denoted as $\pi_{\mathrm{ref}}$) using DPO to obtain the target model $\pi_\theta$. The optimization objective is defined as:
\begin{multline}
\label{DPO} 
\mathcal{L}_{\mathrm{DPO}}\left(\pi_\theta ; \pi_{\mathrm{ref}}\right) = \log \sigma\left( \beta \log \frac{\pi_\theta\left(y_w \mid x,v\right)}{\pi_{\mathrm{ref}}\left(y_w \mid x,v\right)} \right. 
\left. - \beta \log \frac{\pi_\theta\left(y_l \mid x,v\right)}{\pi_{\mathrm{ref}}\left(y_l \mid x,v\right)} \right)
\end{multline}
where $y_w$ and $y_l$ represent the winning ($R_{AB}$) and losing ($R_{\text{AB\_Fake}}$) reports respectively, $\sigma$ is the logistic function, and $\beta$ is a hyperparameter scaling the deviation from the reference model. By training the model to prefer $R_{AB}$ over the clinically confounded $R_{\text{AB\_Fake}}$, we explicitly force the model to capture the subtle visual-textual misalignments. This process directs the model's attention toward fine-grained visual cues that distinguish the two reports, thereby enhancing its discriminative capability and factual consistency.

\section{Experimental details}
\noindent\textbf{Experimental Setting.}

We benchmark our method against existing CT-specific baselines with open-source training code. These include the non-LLM generator CT2Rep \cite{hamamci2024ct2rep}, four CT foundation models, RadFM \cite{wu2023towards}, M3D \cite{bai2024m3d}, CT-CHAT \cite{hamamci2024developing}, and Reg2RG \cite{chen2025large}, a foundation-model-based method tailored for CT. For a fair comparison, all models were fine-tuned on the CT-RATE dataset following their original protocols. As for VideoLMs, we evaluated Qwen2.5-VL \cite{wang2024qwen2} and InternVL3 \cite{zhu2025internvl3} as our backbones. All models were trained on two 80GB NVIDIA A100 GPUs with a total batch size of 4.\\
\noindent\textbf{Dataset.}
% \subsection{Dataset}
% \noindent \textbf{Dataset.} 
We evaluated \textbf{CT-RATE}~\cite{hamamci2024ct2rep} with paired HRCT and paired reports, collected from Istanbul Medipol University. It includes 25,692 noncontrast chest CT volumes from 21,304 unique patients, expanded to 50,188 volumes through various reconstruction techniques. 
% Each volume is paired with corresponding radiology text reports, multiple abnormality labels, and metadata. 
We follow their data splitting method and evaluate our model on their validation set. Each HRCT is converted into 240 fixed slices, each 480 by 480 pixels, within an HU window of [-1000, 200]. To accommodate the video language model, we further save the each CT scan in MP4 format with a frame rate of 18 fps. Following \cite{hamamci2024ct2rep}, we partitioned the dataset into a training set of 46,717 CT scans from 20,000 patients and a validation set of 3,039 CT scans from 1,314 patients.\\

\noindent\textbf{Evaluation metrics.} 

To comprehensively compare the generated reports, we  employ both language-based metrics and clinical-efficacy metrics.
For NLG, BLEU (BL)~\cite{papineni2002bleu}, which measures n-gram overlap with a brevity penalty, emphasizing exact matches; ROUGE-L  (\text{RG\_L})\cite{lin2004rouge}, which evaluates structural similarity using the longest common subsequence; and BERTScore (BERT)\cite{zhang2019bertscore}, which leverages contextual embeddings from BERT to capture deeper semantic alignment between generated and reference texts. For clinical efficacy, we use a pretrained RadBERT classifier~\cite{hamamci2024developing} for classifying 18 abnormalities in radiology reports, The averaged precision, recall and F1 over 18 abnormalities are reported. 

\begin{figure}[t]
    \centering
\includegraphics[width=\linewidth]{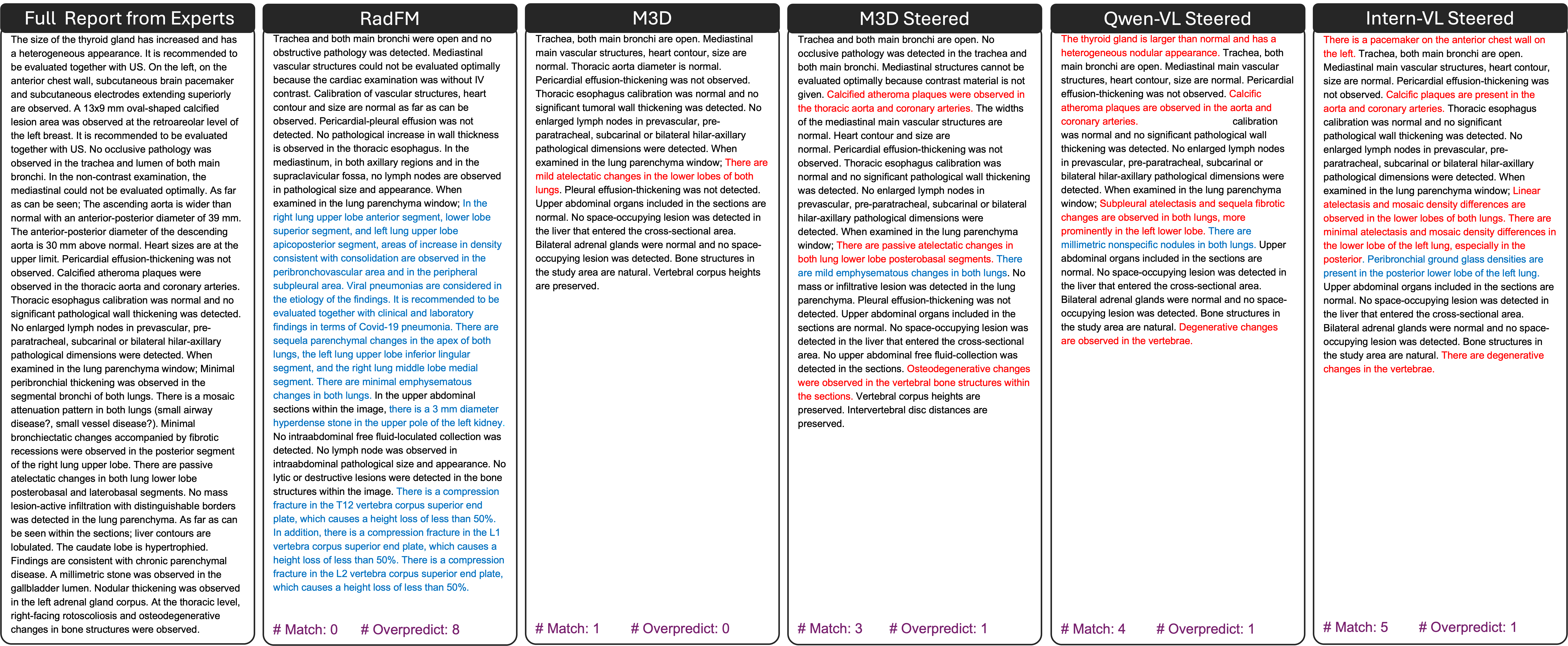}
    \caption{
    Case Study: Comparison of generated reports. Red text denotes ground-truth abnormalities, while green indicates clinical findings absent from the original reports. Total counts for matched and overpredicted findings are provided.
    % Notably, our steered methods identify more clinically significant abnormalities without introducing irrelevant or hallucinated content.
    }
    \label{fig:case_study}
\end{figure}

\section{Experimental results}
\begin{table}[h]
\caption{Benchmark of different methods across various evaluation metrics on the CT-RATE benchmark. All the methods are retrained using CT-RATE, where AbSteered indicate the models finetuned with the proposed approach.}
\label{tab:Benchmark_CTRATE_RAW}
\centering
\resizebox{0.95\textwidth}{!}{
\small 
\begin{tabular}{lcccccccccccccccccccc}
\toprule
 \multicolumn{2}{c}{\textbf{}}  & \multicolumn{7}{c}{\textbf{NLG Metrics}} 
 & \multicolumn{3}{c}{\textbf{CE Micro}} 
 & \multicolumn{3}{c}{\textbf{CE Macro}} 
 & \multicolumn{3}{c}{\textbf{CE Weighted}}  & \multicolumn{3}{c}{\textbf{CE Sample}} \\  
 \cmidrule(lr){3-9} \cmidrule(lr){10-12} \cmidrule(lr){13-15} \cmidrule(lr){16-18} \cmidrule(lr){19-21}
\textbf{Method} & \textbf{Dataset} & \textbf{BL-1} & \textbf{BL-2} & \textbf{BL-3} & \textbf{BL-4} & \textbf{MT-R}  & \textbf{RG-L} 
 
&\textbf{BERT} & \textbf{P} & \textbf{R} & \textbf{F1} & \textbf{P} & \textbf{R} & \textbf{F1} & \textbf{P} & \textbf{R} & \textbf{F1} &\textbf{P} & \textbf{R} & \textbf{F1}
\\
\midrule

CT2Rep \cite{hamamci2024ct2rep} & CT-RATE &
47.91 & \textbf{38.23} & \textbf{32.41} & \textbf{28.04} & \textbf{45.62} & \textbf{45.43} & \textbf{88.10} & 26.39 & 10.50 & 14.10 & 17.03 & 12.59 & 10.65 & 24.44 & 10.37 & 11.35 & 17.36 & 9.07 & 10.86\\

RadFM \cite{wu2023towards}  &  CT-RATE  & \textbf{50.20} & 33.88 & 24.09 & 17.02 & 36.81 & 30.46 & 86.17 & 36.10 & 13.48 & 19.63 & 28.51 & 10.16 & 13.05 & 36.10 & 13.48 & 17.74 & 20.47 & 10.11 & 12.14  \\
Reg2RG \cite{chen2025large}   & CT-RATE
& 44.89 & 33.43 & 26.42 & 21.08 & 38.25 & 24.41 & 86.18 & 28.47 & 11.06 & 15.93 & 19.76 & 8.17  & 10.48 & 26.19 & 11.06 & 14.51 & 19.81 & 11.51 & 12.19\\
CT-CTCHAT  \cite{hamamci2024developing}  & CT-RATE   & 42.81 & 31.07 & 23.38 & 17.63 & 37.62 & 32.50 & 86.35 & 25.13 & 37.48 & 30.08 & 19.97 & 29.68 & 21.66 & 25.70 & 37.48 & 28.35 & 25.25 & 34.35 & 25.31\\
\midrule
M3D-8B \cite{bai2024m3d}  & CT-RATE  &44.95 & 34.36 & 27.83 & 22.98 & 40.75 & 37.76 & 87.52 & 47.60 & 28.54 & 35.69 & 41.62 & 22.41 & 26.74 & 47.06 & 28.54 & 33.13 & 35.04 & 23.21 & 25.21\\

Qwen2.5-VL-7B \cite{wang2024qwen2}   & CT-RATE  & 43.67 & 32.91 & 26.20 & 21.25 & 40.73 & 36.71 & 87.30 & 48.06 & 25.88 & 33.64 & 39.85 & 20.06 & 25.57 & 45.98 & 25.88 & 32.19 & 35.71 & 22.36 & 24.95  \\
InternVL3-8B \cite{zhu2025internvl3}  & CT-RATE  &45.57 & 34.36 & 27.31 & 22.05 & 41.82 & 38.49 & 87.40 & \underline{53.57} & 37.99 & 44.45 & \underline{48.28} & 33.71 & \underline{38.91} & \underline{52.35} & 37.99 & 43.28 & 40.00 & 31.16 & 32.14\\
\midrule

M3D-AbSteer    & CT-RATE & 45.22 & 34.64 & 27.99 & 23.09 & 41.26 & 38.58 & \underline{87.83} & 44.95 & 41.66 & 43.24 & 39.94 & 34.75 & 36.18 & 44.01 & 41.66 & 41.89 & 43.19 & 37.26 & 36.54\\
Qwen2.5-VL-AbSteer    & CT-RATE & 45.64 & 34.14 & 26.81 & 21.40 & 41.61 & 37.99 & 87.13 & 49.15 & \underline{43.22} & \underline{45.99} & 43.13 & \underline{35.78} & 37.90 & 47.05 & \underline{43.22} & \underline{44.05} & \underline{44.86} & \underline{37.58} & \underline{37.39}  \\
InternVL3-AbSteer  & CT-RATE  &\underline{48.32} & \underline{36.66} & \underline{29.17} & \underline{23.58} & \underline{44.13} & \underline{40.49} & 87.59 & \textbf{57.88} & \textbf{51.58} & \textbf{54.55} & \textbf{54.08} & \textbf{45.01} & \textbf{47.66} & \textbf{56.26} & \textbf{51.58} & \textbf{52.80} & \textbf{51.53} & \textbf{45.13} & \textbf{44.80}\\
\bottomrule
\end{tabular}}
\end{table}

\noindent\textbf{Numerical results.} {\textit{VideoLMs exhibit significant potential for high-volume medical report generation, achieving performance comparable to modality-specific foundation models.}}
As illustrated in Table~\ref{tab:Benchmark_CTRATE_RAW}, among the foundation models baseline (above the divider), M3D-8B achieves the leading performance when fine-tuned on the CTRATE dataset. Notably, the general-domain Qwen2.5-VL-7B achieves comparable results to M3D-8B, with only a marginal gap in recall. Furthermore, InternVL3-8B demonstrates even stronger baseline capabilities, occasionally surpassing M3D-8B across key metrics.

\noindent{\textit{Steered general VideoLMs exceed the performance of domain-specific medical foundations.}} As shown in the "AbSteered" section of Table~\ref{tab:Benchmark_CTRATE_RAW}, both the state-of-the-art medical foundation model M3D and the two VideoLM backbones, Qwen2.5-VL and InternVL3, exhibit marked improvements after applying our steering method. Notably, the gains achieved by Qwen2.5-VL and InternVL3 are substantially larger, enabling them to outperform the previously strongest modality-specific report generators by a clear margin. This result not only validates the effectiveness of our abnormality-centric language steering framework, but also suggests that the strong 3D structural reasoning capacity of large-scale VideoLMs can be effectively unlocked through appropriate domain-specific guidance.

\noindent\textbf{Case Study.} A case study is presented in Fig.~\ref{fig:case_study} to illustrate the precision–recall trade-offs across different models in fine-grained abnormality detection.  
As observed, all models benefit significantly from the proposed steering framework. Notably, VideoLM achieves the highest recall among all tested models without increasing hallucinations. \\

\begin{figure}[t]
    \centering
    \begin{minipage}{0.32\linewidth}
        \centering
        \includegraphics[width=\linewidth]{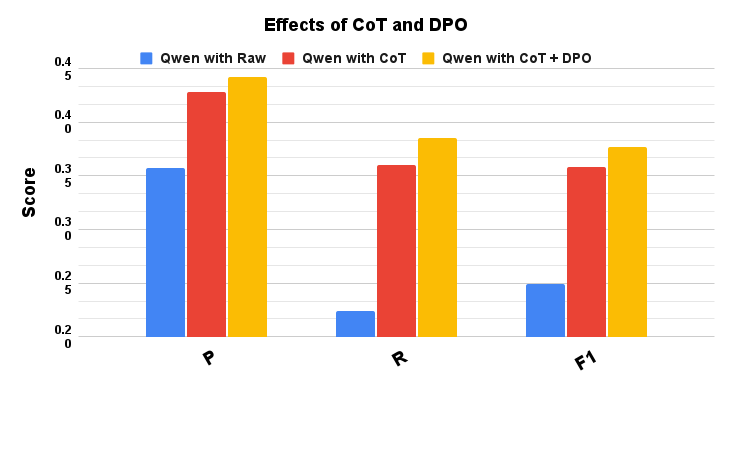}
        % \vspace{-2mm}
        % {\scriptsize (b) Base Model}
    \end{minipage}\hfill
    \begin{minipage}{0.32\linewidth}
    \centering    
    \includegraphics[width=\linewidth]{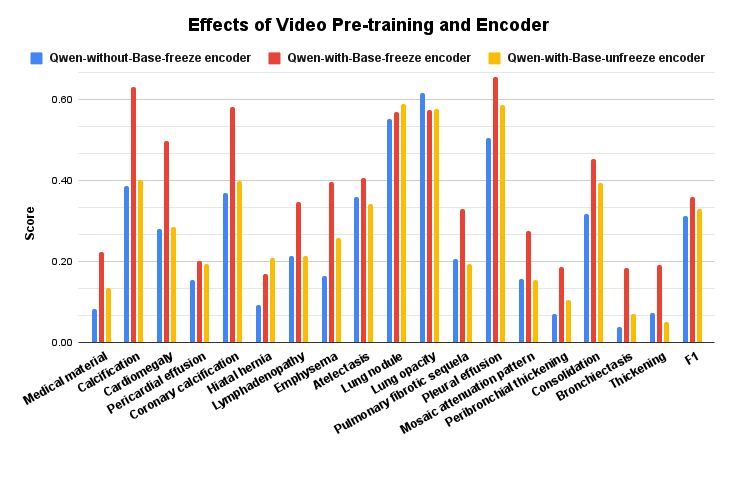}
        % \vspace{-2mm}
        % {\scriptsize (c) Language Steering}
    \end{minipage}\hfill
    \begin{minipage}{0.32\linewidth}
        \centering
    \includegraphics[width=\linewidth]{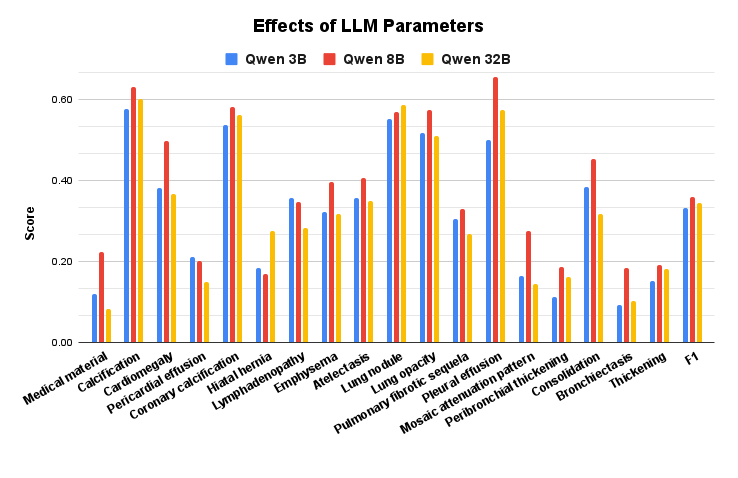}
        % \vspace{-2mm}
        % {\scriptsize (a) LLM Scale}
    \end{minipage}
    \caption{Ablation study on the (a) AbSteering strategy, (b) visual encoder architectures, and (c) LLM parameter scales.}
    \label{fig:ablation_chart}
\end{figure}

\noindent\textbf{Ablation Study.} \noindent\textbf{CoT and DPO.} Fig.~\ref{fig:ablation_chart}(a) shows that abnormality-centric CoT significantly boosts recall. Adding fine-grained DPO further enhances both precision and recall, effectively suppressing hallucinations while increasing sensitivity—a synergy often elusive in conventional methods. \noindent\textbf{Video Pre-training and Encoder.} Using Qwen with Stage 1 CoT as a baseline, we compared: (1) training from scratch, (2) a frozen pre-trained encoder, and (3) Low-Rank Adaptation (LoRA) fine-tuning ($rank=8$). Fig.~\ref{fig:ablation_chart}(b) confirms that discarding pre-trained weights causes sharp performance drops, proving general-domain video pre-training is a vital foundation. Surprisingly, LoRA yields no gains over the frozen encoder, suggesting the pre-trained features are sufficiently robust to generalize to medical textures without further adaptation. \noindent\textbf{LLM Scaling.} We explored 3B, 7B, and 32B LLM scales. Performance improves from 3B to 7B but degrades at 32B (Fig.~\ref{fig:ablation_chart}(c)). This suggests the current bottleneck lies not in LLM capacity, but in visual-textual alignment or data density.

\section{Conclusion}

 HRCT is vital for diagnosing and monitoring thoracic and cardiopulmonary diseases. In this work, we systematically study the transferability of general-domain, video-pretrained vision--language models to HRCT report generation, and quantify the contribution of key components through controlled analyses. We  propose an efficient abnormality-steering framework that effectively adapts these models to domain-specific medical reporting, even with limited in-domain data and specialized clinical knowledge. Overall, our results demonstrate that modern video-pretrained VideoLMs can serve as strong backbones for fine-grained HRCT reporting, and provide an efficient, practical, and data-efficient pathway for repurposing general-domain VideoLMs toward complex, domain-knowledge-intensive medical applications.

\newpage
\bibliographystyle{splncs04}
\bibliography{Main.bbl}

\begin{thebibliography}{10}
\providecommand{\url}[1]{\texttt{#1}}
\providecommand{\urlprefix}{URL }
\providecommand{\doi}[1]{https://doi.org/#1}

\bibitem{achiam2023gpt}
Achiam, J., Adler, S., Agarwal, S., Ahmad, L., Akkaya, I., Aleman, F.L., Almeida, D., Altenschmidt, J., Altman, S., Anadkat, S., et~al.: Gpt-4 technical report. arXiv preprint arXiv:2303.08774  (2023)

\bibitem{bai2024m3d}
Bai, F., Du, Y., Huang, T., Meng, M.Q.H., Zhao, B.: M3d: Advancing 3d medical image analysis with multi-modal large language models. arXiv preprint arXiv:2404.00578  (2024)

\bibitem{cao2024bootstrapping}
Cao, W., Zhang, J., Xia, Y., Mok, T.C., Li, Z., Ye, X., Lu, L., Zheng, J., Tang, Y., Zhang, L.: Bootstrapping chest ct image understanding by distilling knowledge from x-ray expert models. In: Proceedings of the IEEE/CVF Conference on Computer Vision and Pattern Recognition. pp. 11238--11247 (2024)

\bibitem{chen20243d}
Chen, H., Zhao, W., Li, Y., Zhong, T., Wang, Y., Shang, Y., Guo, L., Han, J., Liu, T., Liu, J., et~al.: 3d-ct-gpt: Generating 3d radiology reports through integration of large vision-language models. arXiv preprint arXiv:2409.19330  (2024)

\bibitem{chen2025large}
Chen, Z., Bie, Y., Jin, H., Chen, H.: Large language model with region-guided referring and grounding for ct report generation. IEEE Transactions on Medical Imaging  (2025)

\bibitem{chen2025dia}
Chen, Z., Luo, L., Bie, Y., Chen, H.: Dia-llama: Towards large language model-driven ct report generation. In: International Conference on Medical Image Computing and Computer-Assisted Intervention. pp. 141--151. Springer (2025)

\bibitem{deng2025mvketr}
Deng, X., He, X., Bao, J., Zhou, Y., Cai, S., Cai, C., Chen, Z.: Mvketr: chest ct report generation with multi-view perception and knowledge enhancement. IEEE Journal of Biomedical and Health Informatics  (2025)

\bibitem{hamamci2024developing}
Hamamci, I.E., Er, S., Almas, F., Simsek, A.G., Esirgun, S.N., Dogan, I., Dasdelen, M.F., Durugol, O.F., Wittmann, B., Amiranashvili, T., et~al.: Developing generalist foundation models from a multimodal dataset for 3d computed tomography  (2024)

\bibitem{hamamci2024ct2rep}
Hamamci, I.E., Er, S., Menze, B.: Ct2rep: Automated radiology report generation for 3d medical imaging. In: International Conference on Medical Image Computing and Computer-Assisted Intervention. pp. 476--486. Springer (2024)

\bibitem{hamamci2026generalist}
Hamamci, I.E., Er, S., Wang, C., Almas, F., Simsek, A.G., Esirgun, S.N., Dogan, I., Durugol, O.F., Hou, B., Shit, S., et~al.: Generalist foundation models from a multimodal dataset for 3d computed tomography. Nature Biomedical Engineering pp. 1--19 (2026)

\bibitem{li2025mu}
Li, S., Qin, P., Wu, H., Nie, D., Thirunavukarasu, A.J., Yu, J., Zhang, L.: $\mu$ 2 tokenizer: Differentiable multi-scale multi-modal tokenizer for radiology report generation. In: International Conference on Medical Image Computing and Computer-Assisted Intervention. pp. 3--12. Springer (2025)

\bibitem{lin2023video}
Lin, B., Ye, Y., Zhu, B., Cui, J., Ning, M., Jin, P., Yuan, L.: Video-llava: Learning united visual representation by alignment before projection. arXiv preprint arXiv:2311.10122  (2023)

\bibitem{lin2004rouge}
Lin, C.Y.: Rouge: A package for automatic evaluation of summaries. In: Text summarization branches out. pp. 74--81 (2004)

\bibitem{papineni2002bleu}
Papineni, K., Roukos, S., Ward, T., Zhu, W.J.: Bleu: a method for automatic evaluation of machine translation. In: Proceedings of the 40th annual meeting of the Association for Computational Linguistics. pp. 311--318 (2002)

\bibitem{tang2024work}
Tang, Y., Yang, H., Zhang, L., Yuan, Y.: Work like a doctor: Unifying scan localizer and dynamic generator for automated computed tomography report generation. Expert Systems with Applications  \textbf{237},  121442 (2024)

\bibitem{wang2024qwen2}
Wang, P., Bai, S., Tan, S., Wang, S., Fan, Z., Bai, J., Chen, K., Liu, X., Wang, J., Ge, W., et~al.: Qwen2-vl: Enhancing vision-language model's perception of the world at any resolution. arXiv preprint arXiv:2409.12191  (2024)

\bibitem{wang2024emu3}
Wang, X., Zhang, X., Luo, Z., Sun, Q., Cui, Y., Wang, J., Zhang, F., Wang, Y., Li, Z., Yu, Q., et~al.: Emu3: Next-token prediction is all you need. arXiv preprint arXiv:2409.18869  (2024)

\bibitem{wu2023towards}
Wu, C., Zhang, X., Zhang, Y., Wang, Y., Xie, W.: Towards generalist foundation model for radiology. arXiv preprint arXiv:2308.02463  (2023)

\bibitem{zhang2024vision}
Zhang, J., Huang, J., Jin, S., Lu, S.: Vision-language models for vision tasks: A survey. IEEE Transactions on Pattern Analysis and Machine Intelligence  (2024)

\bibitem{zhang2019bertscore}
Zhang, T., Kishore, V., Wu, F., Weinberger, K.Q., Artzi, Y.: Bertscore: Evaluating text generation with bert. arXiv preprint arXiv:1904.09675  (2019)

\bibitem{zhang2024radgenome}
Zhang, X., Wu, C., Zhao, Z., Lei, J., Zhang, Y., Wang, Y., Xie, W.: Radgenome-chest ct: A grounded vision-language dataset for chest ct analysis. arXiv preprint arXiv:2404.16754  (2024)

\bibitem{zhu2025internvl3}
Zhu, J., Wang, W., Chen, Z., Liu, Z., Ye, S., Gu, L., Tian, H., Duan, Y., Su, W., Shao, J., et~al.: Internvl3: Exploring advanced training and test-time recipes for open-source multimodal models. arXiv preprint arXiv:2504.10479  (2025)

\end{thebibliography}
\end{document}